\documentclass[preprint]{article}

\usepackage{neurips_2020}

\usepackage[utf8]{inputenc} 
\usepackage[T1]{fontenc}    

\usepackage{booktabs}       
\usepackage{amsfonts}       
\usepackage{amsmath}
\usepackage{nicefrac}       
\usepackage{microtype}      
\usepackage{natbib}
\usepackage{graphicx}
\usepackage{todonotes}
\usepackage{layouts}

\usepackage{color}
\definecolor{deepblue}{rgb}{0,0,0.5}
\definecolor{deepred}{rgb}{0.6,0,0}
\definecolor{deepgreen}{rgb}{0,0.5,0}
\definecolor{deeporange}{rgb}{0.6,0.25,0}
\definecolor{verylightgray}{rgb}{0.97,0.97,0.97}
\definecolor{mydarkblue}{rgb}{0,0.08,0.45}
\definecolor{bluegray}{rgb}{0.3,0.3,0.6}

\usepackage{amsthm}
\theoremstyle{plain}            
\newtheorem{thm}{Theorem}[section]

\theoremstyle{definition}       

\newcommand\E{\operatorname{\sf E}}
\newcommand\Var{\operatorname{\sf Var}}

\newcommand\tee{\textsf{\text{T}}\,}

\newcommand{\J}{\operatorname{\sf J}}

\renewcommand{\H}{\operatorname{\sf H}}

\newcommand{\KL}{\operatorname{\sf KL}}

\newcommand{\iid}{\overset{\text{\tiny iid}}{\sim}}
\newcommand{\defeq}{\overset{\text{\tiny def}}{=}}

\usepackage{listings}

\newcommand\pythonstyle{\lstset{
language=Python,
basicstyle=\footnotesize\ttm,
keywordstyle=\ttb\color{deepblue},
commentstyle=\ttm\color{deepgreen},
stringstyle=\color{deeporange},
emphstyle=\ttb\color{deepred},    
emph={MyClass,__init__},          
otherkeywords={self,yield},       
frame=single,                     
showstringspaces=false,
breaklines=true,
backgroundcolor=\color{verylightgray},
}}

\usepackage{enumitem}
\setlist[enumerate]{leftmargin=15pt}  
\setlist[itemize]{leftmargin=15pt}  

\lstnewenvironment{python}[1][]
{
\pythonstyle
\lstset{#1}
}
{}

\newcommand\pythoninline[1]{{\pythonstyle\lstinline!#1!}}

\usepackage{cleveref}

\usepackage[ruled,vlined]{algorithm2e}

\title{Density of States Estimation for \\ Out-of-Distribution Detection}

\author{%
  Warren R.~Morningstar \thanks{Corresponding authors.} \\
  Google Research \\
  \texttt{wmorning@google.com} \\
  \And
  Cusuh Ham \thanks{Work performed during an internship at Google Research.} \\
  Georgia Institute of Technology \\
  \texttt{cusuh@gatech.edu} \\
  \And
  Andrew G.~Gallagher \\
  Google Research \\
  \texttt{agallagher@google.com} \\
  \And
  Balaji Lakshminarayanan \\
  Google Research \\
  \texttt{balajiln@google.com} \\
  \And
  Alexander A.~Alemi \\
  Google Research \\
  \texttt{alemi@google.com} \\
  \And
  Joshua V. Dillon \footnotemark[1]\\
  Google Research \\
  \texttt{jvdillon@google.com}
}
\date{February 2020}

\begin{document}

\maketitle

\begin{abstract}
Perhaps surprisingly, recent studies have shown probabilistic model likelihoods
have poor specificity for out-of-distribution (OOD) detection and often assign
higher likelihoods to OOD data than in-distribution data. To ameliorate this
issue we propose DoSE, the density of states estimator.  Drawing on the
statistical physics notion of ``density of states,'' the DoSE decision rule
avoids direct comparison of model probabilities, and instead utilizes the
``probability of the model probability,'' or indeed the frequency of any
reasonable statistic. The frequency is calculated using nonparametric density
estimators (e.g., KDE and one-class SVM) which measure the typicality of
various model statistics given the training data and from which we can flag
test points with low typicality as anomalous. Unlike many other methods, DoSE
requires neither labeled data nor OOD examples. DoSE is modular and can be
trivially applied to any existing, trained model. We demonstrate DoSE's
state-of-the-art performance against other unsupervised OOD detectors on 
previously established ``hard'' benchmarks.
\end{abstract}

\section{Introduction} \label{sec:intro}
An important assumption behind the success of machine learning methods is that the data seen at inference follows a similar distribution to the training data. When a model encounters an anomalous, or out-of-distribution (OOD) input, it can output incorrect predictions with high confidence. Therefore, it is important to the reliability and safety of these systems to be able to recognize distributional shifts that are often present in real-world applications, such as autonomous driving and medical diagnoses.

The many proposed approaches to OOD detection can be broadly categorized into supervised and unsupervised methods. In a supervised setting, models have access to class labels and/or specific OOD examples, and are either calibrated \textit{post hoc} to flatten the predictive distribution as the distance from the training set increases \citep{liang2018enhancing} or directly trained to distinguish in- and out-of-distribution examples \citep{hendrycks2019deep, meinke2020towards, dhamija2018reducing}.

In an unsupervised setting, generative models are often employed because of their ability to approximate or calculate the density $q(X)$ that describes the distribution of the training set, which can then be used to determine when to trust the prediction $q(Y|X)$. Historically, this approach centers around interpreting this density as a probability of the input $x$, and therefore assuming OOD inputs would be assigned lower probability than in-distribution inputs, making them ``less likely'' to be in-distribution. However, \citet{nalisnick2018do, hendrycks2019deep} exposed some egregious failure modes of this methodology, such as OOD inputs being assigned higher likelihoods than in-distribution examples. Concurrent work by \citet{choi2018waic} and follow-up work from \citet{nalisnick2019detecting} showed that this failure occurs because the typical set of the data may not intersect with the region of high density.

\begin{figure}[!htb]
    \centering
    \includegraphics[width=\textwidth]{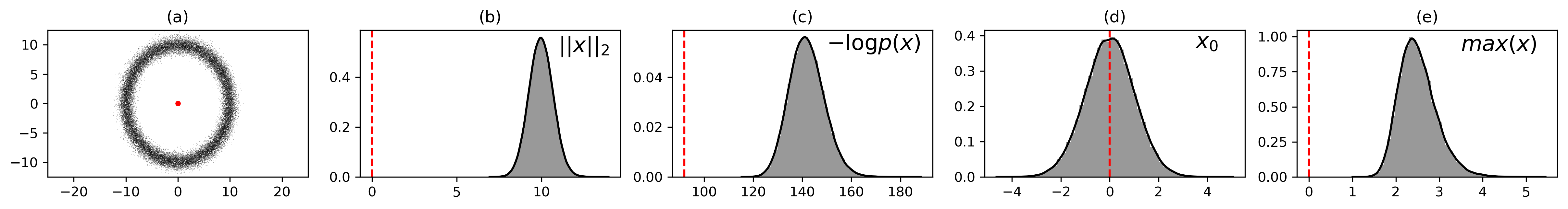}
    \setlength{\abovecaptionskip}{-0.25cm}
    \setlength{\belowcaptionskip}{-0.5cm}
    \caption{(a) A two dimensional projection of a 100 dimensional multivariate normal distribution.  The origin is shown in red. We show histogrammed measurements of 100,000 random draws from this distribution: (b) The observed norm of the draws, (c) the negative log-likelihood, (d) the value of the first coordinate, and (e) the max over the coordinates of the draws. The dashed vertical line denotes the corresponding measurement for the origin.
    }
    \label{fig:gauss}
\end{figure}

 Consider a high-dimensional isotropic Gaussian distribution with zero mean and unit variance. In \Cref{fig:gauss} (a), we show a two-dimensional slice of this distribution. Note that while the mean of this distribution has the highest likelihood (red), it is clearly not \emph{typical} since the likelihood of draws concentrate on lower likelihoods, as shown in (c). This phenomenon is a consequence of the norm's sensitivity to one large variance dimension. The Gaussian Annulus Theorem \citep{blum2020foundations} formalizes this idea and stipulates that samples concentrate on the spherical shell of radius $\sqrt{d}$, as depicted in (b).

 These observations have parallels to physical systems.  In statistical mechanics, the probability of observing a particle in a given state is governed not only by the probability of the state, but also by the geometry of the system.  The density of states codifies this idea; it describes the number of configurations in the system which take on particular values of a given statistic.  \Cref{fig:gauss} (d, e) show how different statistics convey different information about the state.  From this we hypothesize that the density of states--as measured by different statistics--might be potentially useful as a tool for identifying OOD data.

 Our approach for identifying samples as being in- or out-of-distribution is to produce an estimator of this density of states on several summary statistics of the in-distribution data, and then to evaluate the density of states estimator (DoSE) on new trial points, marking those that have low support under the observed densities of the measurements as out-of-distribution.  In general we expect and observe that a relatively small set of reasonable measurements of the samples works well at OOD detection.  We summarize our contributions as follows:

\begin{enumerate}
    \item We propose a novel OOD detection method, DoSE, inspired by ideas from statistical physics and the notion of typicality, that jointly leverages multiple summary statistics from generative models to differentiate between in-distribution and out-of-distribution data.
    \item We show that two variants of DoSE can be easily applied to any pre-trained, generative model. Specifically, we evaluate DoSE with $\beta$-VAEs \citep{higgins2017beta} and Glow \citep{kingma2018glow}.
    \item We evaluate our method on OOD detection benchmarks and demonstrate state-of-the-art performance among unsupervised methods, and comparable performance to state-of-the-art supervised methods.
\end{enumerate}

\section{Related Work}
Given that modern large-scale neural networks can both be fooled by very small perturbations to their inputs~\citep{szegedy2013intriguing}, as well as make poorly-calibrated predictions~\citep{guo2017calibration}, it is increasingly important for neural networks used in applications to be able to identify when it is asked to make predictions for out-of-distribution inputs. 

\textbf{Unsupervised OOD Detection.} \cite{bishop1994novelty} first proposed that generative models may be a useful tool for OOD detection, using a one-sided threshold on the log-likelihood as a decision rule.  The underlying idea behind this approach is that the likelihood represents the ``probability of the data,'' and therefore a high likelihood means that the data is ``good,'' and a low likelihood is ``bad.''  They found that this approach, applied to a model trained on 4 classes was successful at detecting OOD data generated from a fifth class using 16-dimensional feature vectors.  However, the success of this early approach may have been merely coincidental, or due to the fact that the model operated on a low-dimensional feature space.  Notably \cite{choi2018waic} pointed out that in extremely high dimensions, the previously held assumption that in-distribution inputs should have high likelihoods does \emph{not} hold.  This was concurrently validated empirically by \cite{nalisnick2018do, hendrycks2019deep}, who showed that the log-likelihood returned by deep generative models can often be higher for OOD data than it is for in-distribution data. \cite{serra2019input} suggested that the input complexity of the data may be responsible for this effect. 

Subsequent work on unsupervised OOD detection has focused largely on ways to correct this pathology.  For example, \citet{choi2018waic} proposed that OOD data may receive higher likelihoods because of epistemic errors in the likelihood computation, and instead proposed to use the Watanabe-Akaike Information Criterion (WAIC), thereby leveraging multiple generative models trained in parallel to identify OOD data.   Alternatively, \citet{ren2019likelihood} argue that the reason models assign high likelihoods to OOD data is instead because they are confounded by background information present in the dataset.  Thus, they propose to use the likelihood ratio of an autoregressive model trained on in-distribution data with a heavily regularized model trained on mutated pixel data to try to normalize the likelihood by removing contributions from the ``background pixels.''  However, neither of these approaches address the issue with high dimensional likelihoods, and therefore may be unreliable in broader applications.

Even more recently, efforts have been made to attempt to directly measure the typicality of the input data. \citet{nalisnick2019detecting} propose a simple typicality test by flagging a batch of data $X$ if the mean of the generative model log-likelihood ($\log q(\grave{X}|\theta_{n})$) for that batch disagrees with the mean of $q(X|\theta_{n})$ in the training set by a user-specified threshold.  There are two shortcomings to this approach:  First, their test operates on an entire batch, for which all examples are assumed to be either jointly in-distribution or jointly OOD.  Performance noticeably degrades as the batch size decreases to 1. For practical purposes we require a decision rule that can reliably operate on individual samples.  Second, for both VAEs and flow-based models, the likelihood may not be the most informative metric, while its constituents or an alternative might (see \autoref{fig:gauss} or \autoref{sec:metric_composition} in the supplement).

\textbf{Supervised OOD detection.} There are also many proposed approaches to OOD detection that use labeled in-distribution inputs and/or known OOD examples. \citet{alemi2018uncertainty} show that metrics like \emph{rate} in a Variational Information Bottleneck (VIB) \citep{alemi2017deep} model have reasonable performance on outlier detection and adversarial defense.  ODIN \citep{liang2018enhancing} uses a combination of techniques to calibrate the predictions on in- and out-of-distribution examples from pre-trained deterministic models. \citet{hsu2020generalized} generalize ODIN to not require OOD examples, but still leverage supervised classifiers. \citet{lakshminarayanan2017simple} propose to train ensembles of classifiers to calibrate predictions and soften confidence estimates where the models do not agree. \citet{hendrycks2019deep} use specific OOD examples fed to the model to encourage OOD inputs to have uniform confidence, and therefore be identifiable. \citet{stutz2019confidence} use adversarial attacks to lower the confidence of inputs in an $\epsilon$-ball around the training data. \citet{meinke2020towards} propose to fit a mixture of Gaussians to the latent space distribution of a trained model for both the in-distribution data, and a user-specified OOD dataset, and use the likelihood ratio of the two to lower confidence estimates for points far from the latent space occupied by the training data.  All of these methods have demonstrated successful performance, but are trained with either class labels, or specific outlier examples.

In this work we do not use class labels or any exposure of OOD data to the model during training.  This presents a significant practical advantage over these methods for several reasons:  First, in many settings one may need to identify OOD data without being given class labels.  Second, training specifically to predict class labels may otherwise discard information that may be useful when identifying OOD data (though it also may highlight information which \emph{is} useful).  Third, models trained using specific instances of OOD data are overly tuned to attributes in the OOD dataset, and therefore may suffer from overconfident yet incorrect predictions when given inputs from a separate OOD set.

\section{Approach}\label{sec:Approach}

%
%
%
%
%
%

We first establish notation. Assume access to data generated according to $\{X_i=x_i\}_i^n \iid p(X)$ for $X\in\mathcal{X}$
and that our task is to construct a summary statistic $T_n$ suitable for evaluation on unseen data.  Example summary statistics include $T_n^{(\text{nll})}(X) \defeq -\log q(X|\theta_n)$, $T_n^{(L_2)}(X) \defeq \|X - \mu_n\|_2$, or $T^{(\text{ml})}(\{X,Y\}) \defeq \max_\mathcal{Y}
q(Y|X,\theta_n)$.  Suppose however, that each unseen sample datum is drawn
from the mixture $\grave{X} \sim \alpha p(\grave{X}) + (1-\alpha)\tilde{p}(\grave{X})$
where $\alpha$ and $\tilde{p}$ are unfixed and unknown confounders and
$\tilde{X}\sim\tilde{p}$ has $\tilde{X}\in\mathcal{X}$. Our task is to devise a decision rule for identifying when $T_n(\grave{X})$ is not to be trusted.

Since we presume $\alpha,\tilde{p}$ are unfixed and unknown we can neither access OOD samples $\{\tilde{X}_i\}_i^m$ nor make assumptions of $\alpha,\tilde{p}$ when constructing $T_n$. Our only option is to devise a rule based solely on $T_n$ and $\{x_i\}_i^n$. Our proposal--and indeed an obvious idea--is to fit a distribution to $\{T(x_i)\}_i^n$ and use that probability as a threshold for classifying whether a sample is OOD. For example, assuming the statistic $T$ is multivariate ($T:\mathcal{X}\to \mathbb{R}^D$) one could use a product-of-experts (POE) kernel density estimator (KDE) of the form,
\begin{equation}
    q(X=x|T, \{x_i\}_i^n, h) = \prod_d^D \frac{1}{nh_d} \sum_i^n  \phi_j\left(\frac{[T(x)]_d - [T(x_i)]_d}{h_d}\right),
\end{equation}
one-class SVM \citep{scholkopf2000support}, or any other similarly constructed density.

\subsection{What is a good $q$ in theory?}
What makes for a good OOD distribution, $q$?
How do we choose the statistic $T$ and associated density estimator hyperparameters when we have neither $\tilde{p}$ nor samples from it? Similarly, why is direct use of the maximum likelihood distribution a generally poor OOD detector? \citep{nalisnick2018do}

To answer these questions, consider a slightly generalized notion of the information theoretic typical set,
\begin{equation}\label{eq:pqtypical}
  \mathcal{A}_{p,q}^{(s,\epsilon)}
  = \left\{\{X_i\}_i^s\in\mathcal{X}_p^s : \left|-\frac{1}{s}\sum_i^s\log q(X_i) - \H[p] \right| \le \epsilon \right\},
\end{equation}
where $\H[p]$ is the entropy of the true process $p$, $\epsilon$ governs the permissible entropic gap, $s$ is the sequence length, and $q$ is any
distribution over $\mathcal{X}_p$. Equation~\ref{eq:pqtypical} generalizes the standard definition, $\mathcal{A}_{p,p}^{(s,\epsilon)}$ \citep{coverandthomas} by considering the typicality coverage on $p$ by a possibly different distribution $q$. To make an effective OOD classifier, we are concerned with identifying the $q$ which maximizes the expected typicality of $q$ on $p$, i.e., $\max_{q\in\mathcal{Q}} p(X^s \in \mathcal{A}_{p,q}^{(s,\epsilon)})$. Notably, our work is concerned with the $s=1$ case, i.e., capability for singleton OOD designation.
Theorem~\ref{thm:stat_bias_var} clarifies this objective by way of bound.

\begin{thm}{Bias/Variance Tradeoff for Typicality.}\label{thm:stat_bias_var}
\begin{align}
p(\{X_i\}_i^s\not\in\mathcal{A}_{p,q}^{(s,\epsilon)}) \epsilon^2 \le \KL[p,q]^2 + \tfrac{1}{s}\Var_p[\log q(X)]
\end{align}
\begin{proof}
Write $Y=-\tfrac{1}{s}\sum_i^s \log q(X_i) - \H[p]$. From Markov's inequality, $p(|Y|\ge \epsilon)\epsilon^2 \le \E_p[Y^2]$. Making substitutions based on $\tfrac{1}{t^2} \sum_i^s \Var_p[\log q(X_1)] = \Var_p[\tfrac{1}{s}\sum_i^s \log q(X_i)] = 
\E_p[(-\tfrac{1}{s} \sum_i^s \log q(X_i))^2] - \H[p,q]^2$
and  $\KL[p,q]^2=(\H[p,q] - \H[p])^2$ completes the proof.
\end{proof}
\end{thm}

Through the lens of Theorem~\ref{thm:stat_bias_var}, we understand the MLE-fitted distribution's shortcomings as an OOD probability measure. When $q$ is chosen solely to minimize $\KL[p,q]$, it will generally be a looser bound on the $s=1$ typical set--the case of interest when making single sample OOD evaluations. Likewise, many choices of $T$ are also apparently sub-optimal.  For example, $T^{(42)}(x)=42$ would generally be useless because any density $q$ built solely from this statistic would have an infinite $\KL[p,q]$ unless $p(X)=\delta(X-42)$. Also, we can generally rule out degenerate KDEs ($h=0$) because of their lack of smoothness, i.e., disregard for $\Var_p[\log q]$. 

\subsection{What is a good $q$ in practice?}\label{sec:Heuristics}

Although Theorem~\ref{thm:stat_bias_var} is cognitively appealing, it is not directly computable owing to its nonlinear dependency on $\H[p]$ (an unknown). We now describe a heuristic procedure for minimizing the right-hand side of Theorem~\ref{thm:stat_bias_var} and justify this procedure both by exploring a plugin estimate to Theorem~\ref{thm:stat_bias_var} and by appealing to rationale from statistical physics.

We first note that it is possible to make coarse tuning to the OOD detecting distribution $q$ via crude approximations to \ref{thm:stat_bias_var}'s implications. The empirical approximation of the right-hand of Theorem~\ref{thm:stat_bias_var} over the held-out distribution $\{x_i\}_i^m$ is,
\begin{equation}\label{eq:mcbiasvar}
\KL[p,q]^2 + \Var_p[\log q] \approx \frac{1}{m}\sum_j^m (\log q(x_j|\{x_i\}_i^n,T,\gamma))^2 + 
2\H[p] \frac{1}{m}\sum_j^m \log q(x_j|\{x_i\}_i^n,T,\gamma) + c
\end{equation}
where $m$ is the size of the evaluation set, $\gamma$ represents the parameters of our density (e.g., $h$ for a KDE and $\nu$ for one-class SVM), and $c$ is constant for any choice of $q$. A general strategy to minimize \ref{eq:mcbiasvar} is to consider several different choices for $\H[p]$ and explore different choices of $T$ under this range. Alternatively, one can consider using $\H[q(X|{\theta_n^{(\text{ml})}})]$ as a plugin estimate for $\H[p]$.  This is the “resubstitution estimator” introduced by \cite{beirlant1997nonparametric} and used by \cite{nalisnick2019detecting}. Assuming the OOD distribution has the same discrete support, one can additionally explore use of entropy bounds like $\H[p] \in [0, h\cdot w \cdot c \cdot \log k]$ (for image height $h$, width $w$, channels $c$, and discrete pixel intensity levels $k$) or use or use known estimates of $\H[p]$, e.g., \citet{parmar2018image}. We emphasize that discrete entropy is only reasonable if $q$ also has the same support; failing this requirement may introduce an inconsistent sign in Equation~\ref{eq:mcbiasvar}.

We use equation~\ref{eq:mcbiasvar} and the resubstitution estimator for the entropy to evaluate the tightness of the bound for all different statistics.  For a statistic which is completely informative about the typicality (i.e., it minimizes the bound from \autoref{thm:stat_bias_var}), one need only evaluate that statistic to evaluate the typicality of trial points and identify those which are out-of-distribution.  In practice we find that multiple different statistics get indistinguishable values for this bound, and therefore we do not know which statistic is the most informative.   We therefore construct our estimator based on the KDE of multiple different statistics evaluated on the same data.  The procedure is straightforward:  If we interpret the KDE estimates as probabilities of typicality, then the product of the KDEs gives the probability that a given input is typical for all metrics jointly (assuming no correlation between statistics).  We can further relax the assumption of independence by jointly evaluating the DoSE using an alternative density estimator, such as a one-class Support Vector Machine \citep[SVM; ][]{schoelkopf2001estimating} in our case.  We show in the experiments that both of these approaches outperform alternative methods, which only query a single statistic.  We further show in \autoref{appendix:degradation}, that both of these approaches are robust against the inclusion of uninformative, or even  obfuscatory statistics.

Our procedure to construct DoSE for OOD detection is as follows:
\begin{enumerate}
\item Train a deep probabilistic model $q(X|\theta_n)$ using training set $\{x_i\}_i^n$ where $n$ is the size of a training set from which $m$ samples (chosen randomly) are excluded from training and used as a validation set.
\item Evaluate summary statistics $T_n(x)$ on the training data.  
\item Construct DoSE using a KDE or SVM on each set of statistics from the training set.
\item Evaluate the DoSE score by computing the sum of the log-probabilities from the KDE on each statistic for each example in the training set $\{x_i\}_i^n$ and validation set $\{x_i\}_i^m$.  Alternatively, compute the scores for both sets using the SVM.
\item Check the DoSE calibration between the training and validation DoSE scores using the expected calibration error \citep[ECE; ][]{guo2017calibration}.
\item Determine threshold for OOD rejection, by choosing a number of examples to discard from the validation set, and identifying the corresponding threshold to place on the DoSE score. 
\end{enumerate}

We now establish intuition to further explain the underpinnings of our empirical methodology. In statistical physics, a system contains particles $x$.  For each particle, a measurement or statistic $T_n$ represents a physical property of that particle.  Our challenge is to determine if any given particle is atypical, using only the physical properties of that particle, along with the physical properties of $n$ particles from the system.  Atypicality here means that the particle should not be found having these properties assuming that the system is in equilibrium (i.e. the particle is an anomaly).  For any physical property (e.g., energy), the probability of occurrence in a physical system is determined by the \emph{density of states}: $g(T)=\int dX \delta(T^{\prime}(X)-T)$.  This quantity describes the number of occupied configurations in the system which have a given value of $T$.  

One can often calculate the statistical physics notion of density of states from first principles.  Since this is not possible in our problem setup, we instead simply approximate the density of states using a \emph{Density of States Estimator} (DoSE): a nonparametric density estimator trained to measure the density of states of a statistic $T$ evaluated on an input $\grave{X}$ using the sample particles from the system.  We can apply this approach towards any statistic $T$ evaluated on the data $\{x_i\}_i^n$ to construct the DoSE of that statistic.  DoSE then measures the empirical density of the statistic $T$ evaluated at some new point $\grave{x}$ using nearby points in the training set.  Note that we need not offer any interpretation for $T$, and even if the statistic is not interpretable, we can still measure its typicality.  

\section{Experimental Setup}
We now evaluate the empirical performance of DoSE, following the procedures outlined in \cite{choi2018waic,nalisnick2019detecting,ren2019likelihood}.  To summarize, we first train an ensemble of deep generative models on a given in-distribution dataset.  We then evaluate statistics on examples from the training set and construct our DoSEs using the measured statistics.  We validate that our models are not memorizing using a heldout set of examples from the training set.  We finally compute the DoSE scores on the in-distribution test set, and several OOD datasets.  We measure the success of OOD identification using the Area Under the ROC Curve (AUROC).

We compare our performance against several established unsupervised baselines:
\begin{enumerate}
\item A single-sided threshold on the log-likelihood $q(X|\theta_n)$ \citep{bishop1994novelty}.
\item The \emph{single-sample} typicality test (TT) from \cite{nalisnick2019detecting}.  To evaluate the AUROC using this method, we simply use the raw typicality score $\textrm{TT}(\grave{X})=| \log q(\grave{X}|\theta_n)-\H\left[q(X|\theta_n)\right]|$.  Similar to Nalisnick et al. (2019), we calculate $\H\left[q(X|\theta_n)\right]$ as an empirical average over the training set.

\item The \emph{Watanabe-Akaike Information Criterion} (WAIC) from \cite{choi2018waic}.  For this, we use 5 models trained separately and measure $\textrm{WAIC}(\grave{X}) = \E_{\theta}[\log q(\grave{X}|\theta_n)] - \Var_{\theta}[\log q(\grave{X}|\theta_n)]$

\item  The likelihood ratio method (LLR) from \cite{ren2019likelihood}.  To compute LLR, we train a background model using their proposed method of mutations, using a mutation rate of 0.15, the center of the range in which they found successful results.  The LLR score is then simply $\textrm{LLR}(\grave{X}) = \log q_{s}(\grave{X}|\theta_n) - \log q_{b}(\grave{X}|\theta_n)$, where the subscripts $s$ and $b$ indicate the semantic and background models, respectively.
\end{enumerate}

For all of these methods, we use the \emph{same} models to evaluate the OOD scores.  This highlights the difference in performance caused by the methodology, rather than due to differences in the training procedure.  To quantify the uncertainty in performance resulting from the parameters $\theta$ found during an individual training run, we train 5 separate models in parallel, and evaluate the performance of \emph{all} methods using all models.

For DoSE on $\beta$-VAEs, we used 5 statistics:
(1) posterior/prior cross-entropy, $T^{(\text{xent})}_{n}(X)=\H[q(Z|X,\theta_n), q(Z)]$,
(2) posterior entropy, $T^{(\text{ent})}_{n}(X)=\H[q(Z|X,\theta_n)]$, 
(3) posterior/prior KL divergence, $T^{(\text{rate})}_{n}(X)=\KL[q(Z|X,\theta_n), q(Z)]$,
(4) posterior expected log-likelihood, $T^{(\text{distortion})}_{n}(X)=\E_{q(Z|X,\theta_n)}[\log q(X|Z,\theta_n)]$, and
(5) IWAE \citep{burda2015importance}, $T^{(\text{iwae})}_{n}(X)=\log \E_{q(Z|X,\theta_n)}[q(X,Z,\theta_n)/q(Z|X,\theta_n)]$.
In all cases, the intractable expectation $\E_{q(Z|X,\theta_n)}[f(Z)]$ was replaced with a seeded Monte Carlo approximation, $\tfrac{1}{16}\sum_t^{16}f(Z_t)$ with $Z_t \iid q_\text{post}(Z|X,\theta_n,\text{seed=hash}(X,t))$. By seeding, we ensure the statistics' reproducibility yet preserve the logic of the approximation.
For Glow models, we used 3 metrics: (a) the log-likelihood $T^{(\text{like})}_{n}(X)=q(X|\theta_n)$, (b) the log-probability of the latent variable $T^{(\text{latent})}_{n}(X)=q(Z|X,\theta_n)$, and (c) the log-determinant of the Jacobian between $X$; the input space,  and $Z$; the transformed space (i.e., $T^{(\text{jac})}_{n}(X)\defeq\log |\J(X)|$). 
Additional model and training details are in  \Cref{appendix:experiments} in the supplement.

\section{Results}

A summary of all quantitative results on all baselines is presented in \autoref{tab:results}.  We show the AUROC computed between all pairs of in- and out-of-distribution data, measured using our method as well as alternative techniques.

\begin{table}[!htb]
\fontsize{8pt}{8pt}\selectfont
\begin{center}
\begin{tabular}{llllllll}
Dataset/OOD Dataset & Model & $q(X|\theta_n)$ & WAIC & TT & LLR & DoSE$_\text{KDE}$ & DoSE$_\text{SVM}$ \\ 
\hline
MNIST & VAE \\ 
\hline
Omniglot & & 1.000 & 1.000 & 1.000 & 0.470 & 1.000 & \bf{1.000} \\
FashionMNIST & & 0.998 & 0.988 & 0.997 & 0.404 & \bf{0.999} & 0.996 \\
Uniform & & 1.000 & 1.000 &  1.000 & 0.277 & \bf{1.000} & \bf{1.000} \\
Gaussian & & 1.000 & 1.000 & 1.000 & 0.228 & \bf{1.000} & \bf{1.000} \\
HFlip & & 0.839 & \bf{0.861} & 0.776 & 0.473 & 0.760 & 0.812 \\
VFlip & & \bf{0.838} & 0.821 & 0.837 & 0.499 & 0.818 & 0.830 \\
\hline
FashionMNIST & VAE \\
\hline
Omniglot & & 0.995 & 0.893 & 0.991 & 0.508 & \bf{1.000} & 0.998 \\
MNIST & & 0.931 & 0.950 & 0.901 & 0.503 & \bf{0.998} & 0.997 \\
Uniform & & 0.998 & 0.878 & 0.998 & 0.573 & \bf{1.000} & 0.998 \\
Gaussian & & 0.997 & 0.852 & 0.997 & 0.501 & \bf{1.000} & 0.998 \\
HFlip & & 0.658 & 0.503 & 0.599 & 0.479 & \bf{0.658} & 0.625 \\
VFlip & & 0.702 & 0.473 & 0.635 & 0.485 & \bf{0.748} & 0.728 \\
\hline
CIFAR10 & Glow \\
\hline
CIFAR100 & & 0.520 & 0.532 & 0.548 & 0.520 & 0.569 & \bf{0.571} \\ 
CelebA & & 0.914 & 0.928 & 0.848 & 0.914 & 0.976 & \bf{0.995} \\
SVHN & & 0.064 & 0.143 & 0.870 & 0.064 & \bf{0.973} & 0.955 \\
ImageNet32 & & 0.794 & 0.870 & 0.754 & 0.795 & 0.914 & \bf{0.930} \\
Uniform & & 1.000 & 1.000 & 1.000 & 1.000 & \bf{1.000} & \bf{1.000} \\
Gaussian & & 1.000 & 1.000 & 1.000 & 1.000 & \bf{1.000} & \bf{1.000} \\
HFlip & & 0.501 & 0.499 & 0.500 & 0.501 & \bf{0.507} & 0.502 \\
VFlip & & 0.505 & 0.505 & 0.501 & 0.505 & \bf{0.533} & 0.523 \\
\hline
SVHN & Glow \\
\hline
CelebA & & 1.000 & 0.991 & 1.000 & 0.912 &  \bf{1.000} & \bf{1.000} \\
CIFAR10 & & \bf{0.990} & 0.802 & 0.970 & 0.819 & 0.988 & 0.962 \\
CIFAR100 & & \bf{0.989} & 0.831 & 0.965 & 0.779 & 0.986 & 0.965 \\
ImageNet32 & & 0.998 & 0.980 & 0.994 & 0.916 & 0.999 & \bf{0.999} \\
Uniform & & 1.000 & 1.000 & 1.000 & 1.000 & \bf{1.000} & \bf{1.000} \\
Gaussian & & 1.000 & 1.000 & 1.000 & 1.000 & \bf{1.000} & \bf{1.000} \\
HFlip & & 0.504 & 0.502 & 0.499 & 0.502 & \bf{0.520} & 0.512 \\
VFlip & & 0.502 & 0.504 & 0.500 & 0.501 & 0.510 & \bf{0.511}\\
\hline
CelebA & Glow \\
\hline
CIFAR10 & & 0.404 & 0.507 & 0.634 & 0.323 & 0.861 & \bf{0.949}\\
CIFAR100 & & 0.427 & 0.535 & 0.671 & 0.357 & 0.867 & \bf{0.956} \\
SVHN & & 0.008 & 0.139 & 0.982 & 0.028 & 0.993 & \bf{0.997} \\
ImageNet32 & & 0.705 & 0.837 & 0.775 & 0.596 & 0.995 & \bf{0.998} \\
Uniform & & 1.000 & 0.961 & 1.000 & 1.000 & \bf{1.000} & \bf{1.000} \\
Gaussian & & 1.000 & 1.000 & 1.000 & 1.000 & \bf{1.000} & \bf{1.000} \\
HFlip & & 0.600 & 0.754 & 0.526 & 0.529 & 0.945 & \bf{0.985} \\
Vflip & & 0.706 & 0.734 & 0.602 & 0.606 & 0.983 & \bf{0.998} \\

\end{tabular}
\setlength{\belowcaptionskip}{-0.5cm}
\caption{A comparison of AUROC of our method against unsupervised baselines on the OOD detection task. We find that our method most reliably achieves SoTA performance across all datasets.}\label{tab:results}
\end{center}
\end{table}

We find that, for all ``hard'' dataset pairings, both variants of DoSE either outperform or significantly outperform all competing methods.  Note that this same result is observed for either DoSE evaluated on an individual model or on a full ensemble of models.  For an individual model, we observe that all 5 runs of DoSE$_\text{KDE}$ outperform all 5 runs of all competing techniques.  This corresponds to a probability of $0.003$ that our result was observed due to random chance, compared against any competing technique.  We further find that our method generally outperforms competing techniques on most easy dataset pairings as well, with a few exceptions (e.g., SVHN$\rightarrow$CIFAR10), which are typically found by a one-sided threshold on the likelihood $q(X|\theta_n)$.  While DoSE may not then be the highest performing technique in all dataset pairings, it is important to note that it is the highest performing overall, with an average ranking of 1.2 for both DoSE$_\text{KDE}$ and DoSE$_\text{SVM}$ against other competing techniques (we exclude the other when computing the ranking).  For reference, $q(X|\theta_n)$ has rank of 2.2, TT 2.7, WAIC 3.06, and LLR 4.19.  

\begin{figure}
    \centering
    \includegraphics[width=0.8\hsize]{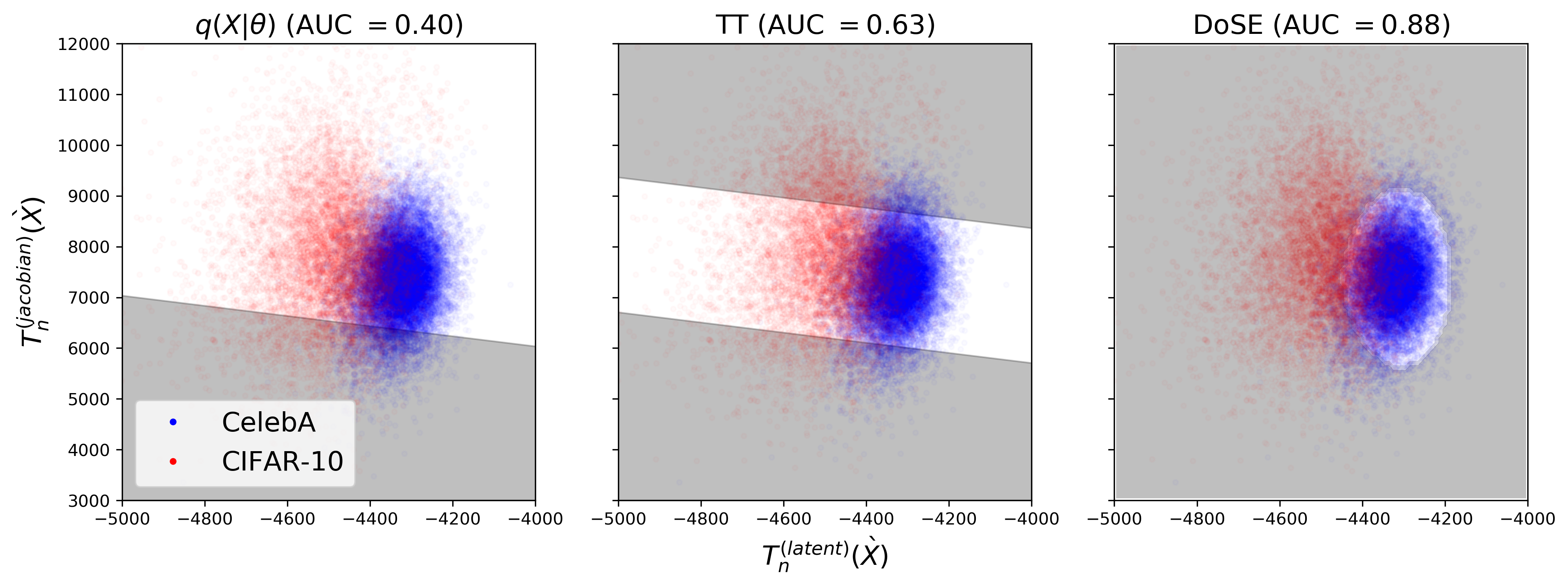}
    \caption{Decomposition of $q(X|\theta_n)$ for a Glow model trained on CelebA.  The blue points show the test data, in coordinates of $T^{(\text{latent})}_{n}(\grave{X})$ and $T^{(\text{jac})}_{n}(\grave{X})$.  Red points show the same coordinates observed for CIFAR10, an OOD dataset.  We show the decision boundaries that exclude 10\% of the in-distribution data for $q(X|\theta_n)$ (left), TT (middle) and DoSE (right).  The shaded region is classified as out-of-distribution, and the non-shaded region is classified as in-distribution.} 
    \label{fig:statistic_space}
\end{figure}

In general, we find that TT achieves more reliable performance than the alternatives.  This is, on some level, to be expected because TT also attempts to directly measure the typicality of a datum.  However, we also find several situations where TT is vulnerable because it relies exclusively on the likelihood.  In particular, we find that TT measures only AUROC $\approx 0.6$ when trying to identify CIFAR10 or CIFAR100 when trained on CelebA.  This result was also observed by \cite{nalisnick2019detecting}, who attributed it to a ``fundamental limitation of deep generative models.''  However, we observe that this is simply due to the fact that the log-likelihood is itself a sum of two different statistics.  Projected into the space of these statistics, CIFAR10 is easily identified as OOD.  We show this decomposition in \autoref{fig:statistic_space}.  Because DoSE uses both of these statistics to identify OOD data, it does not suffer from this vulnerability.  

We find that $q(X|\theta_n)$, WAIC, and LLR all exhibit performance that is much less consistent for different dataset pairings.  In part, we attribute this to the fact that none of these methods attempt to measure the typicality of an input, and are therefore vulnerable to OOD datasets which are assigned anomalously high likelihoods.  As such, all of these methods fail on CIFAR10$\rightarrow$SVHN, CelebA$\rightarrow$CIFARs, CelebA$\rightarrow$SVHN.  For LLR, we may also violate the implicit assumptions underlying the methodology by using models such as VAEs, which may not be able to explicitly decompose the likelihood into semantic and background components in the same way autoregressive models do.  We therefore speculate that LLR may be more successful if a different model were used, though we also note that it still would not measure typicality.

\begin{figure}[ht!]
    \centering
\includegraphics[width=\linewidth]{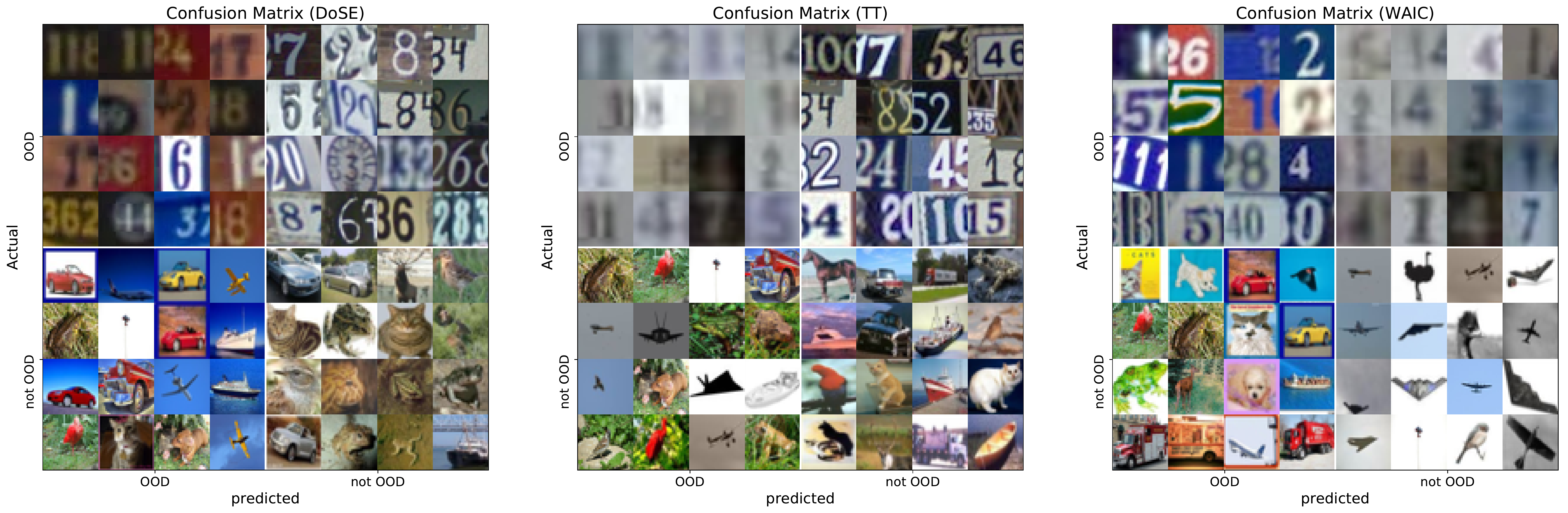}
    \setlength{\abovecaptionskip}{-0.5cm}
    \setlength{\belowcaptionskip}{-0.25cm}
    \caption{Confusion matrices for methods used in our OOD detection performance.  The images in each quadrant of the matrix are in raster order, sorted by the confidence of the classification.}  
    \label{fig:qualitative}
\end{figure}

We also perform a qualitative examination of the attributes in the data which appear to be most indicative of the OOD score from each method.  To do so, we take the 16 images with the highest and lowest OOD scores from a given in- and out-of-distribution pairing for a given method.  These images correspond to the 4 elements of the confusion matrix for each method.   We organize these images into their respective category in the confusion matrix, and show the results for TT, WAIC and DoSE in Figure~\ref{fig:qualitative} on CIFAR-10$\rightarrow$SVHN.  While it is difficult to provide an entirely objective assessment of these results, we speculate that DoSE identifies images with high color contrast as likely OOD candidates.  TT appears to identify a split between images with uniform backgrounds, and images with noisy backgrounds as false positives.  This makes sense, given that these are the images with highest and lowest log-evidence, respectively.  WAIC appears to identify images with irregular colors as likely OOD candidates.  Of course, despite its reasonable qualitative results, WAIC also gets 0.06 AUROC on this particular dataset pairing, undermining its utility as an OOD detection method.

\section{Conclusion}
We have presented a novel method, DoSE, for detecting out-of-distribution data, which can be easily applied to any pre-trained generative model or ensemble of generative models without any additional tuning or modification.  We show that this approach is advantageous over likelihood-based approaches because it provides multiple ways of evaluating the typicality of an input under the assumed generative model.  DoSE does not require class labels or access to specific OOD examples. Leveraging the argument that likelihoods should not be interpreted as the probability that an input is in- or out-of-distribution as well as ideas from statistical physics, our method uses nonparametric density estimators to directly measure the typicality of various model statistics given the training data. We demonstrated state-of-the-art performance with DoSE among unsupervised methods on common OOD detection benchmarks. 

\section*{Broader Impact}
Safe deployment of statistical models necessitates detection of test points which sufficiently deviate from modeling assumptions. Failure to do so risks model outputs which are incorrect and/or meaningless and could result in misguided or dangerous decisions/actions/policies. General purpose techniques which alert model users of this circumstance therefore have immediately positive social impact. In situations where model fairness is a concern, effective OOD detection could identify misrepresented demographic categories. In life-and-death models, e.g., self-driving cars and medical diagnosis, OOD detection could mandate human intervention and potentially prevent catastrophe.


\bibliographystyle{plainnat}
\bibliography{references}

\clearpage
\newpage
\appendix

\section{Isotropic Gaussian Densities}
\label{appendix:isotropic}

Here we work through the simple example given in the main text in 
detail. 

A high-dimensional spherically symmetric Gaussian distribution with mean zero and unit variance in $D$ dimensions has the probability density:
\begin{equation}
    p(X=x)\, dx = \frac{1}{(2\pi)^{D/2}} \exp \left( - \frac{x^\tee x }{2} \right) \, dx.
\end{equation}
Transformed for spherical coordinates, this becomes a
distribution over the norm of the vectors:
\begin{equation}
    p(R=r) \, dr = \frac{2 r^{D-1}}{2^{\frac D 2} \Gamma \left( \frac{D}{2} \right) } \exp \left( -\frac{r^2}{2} \right) \, dr
\end{equation}

The \emph{energy} of the original distribution is:
\begin{equation}
    u \defeq -\log p(X=x) = -\frac{x^\tee x}{2} - \frac D 2 \log (2 \pi) = -\frac{r^2}{2} - \frac{D}{2} \log (2\pi)
\end{equation}

The density of states in this case is given by:
\begin{equation}
    p(u) = \frac{(2\pi)^{\frac D 2}}{\Gamma\left(\frac D 2 \right)} e^{-u} \left( u - \frac D 2 \log 2\pi \right)^{\frac D 2 - 1} 
\end{equation}

\section{Vulnerability of Likelihoods in Flow-based Models}
\label{sec:metric_composition}
\begin{figure}[h]
    \centering
    \includegraphics[width=\hsize]{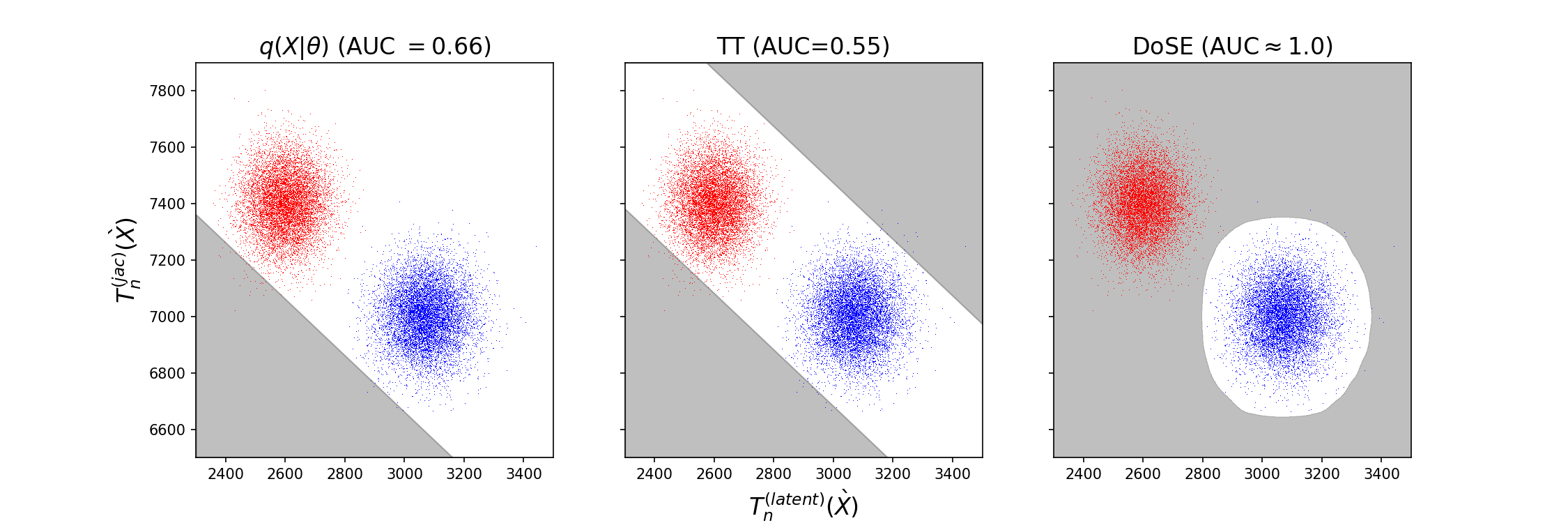}
    \caption{In this toy example, we show the distribution of two statistics, $\log q(Z)$ and $\log{|\J|}$, returned from a flow-based model on in-distribution data (blue) and OOD data (red). Each panel shows the decision regions produced by different OOD detection techniques operating on these metrics.  The left column shows the decision boundaries produced using the log-likelihood.  The middle column shows the decision boundaries produced by TT, a typicality test of the log-likelihood.  The right column shows the decision boundaries produced by DoSE.  In this particular case, the likelihood is the least useful projection over which to attempt to identify OOD data, leading to poor performance of both TT and $q(X|\theta_n)$.  DoSE achieves approximately perfect OOD detection in this same setting.}
    \label{fig:metric_space}
\end{figure}

In many previous works on unsupervised OOD detection \citep[e.g., ][]{nalisnick2019detecting, ren2019likelihood,choi2018waic,bishop1994novelty}, it has been taken for granted that the likelihood $q(X|\theta_n)$ (which is usually the optimization target for a deep generative model) should be the most informative statistic either by interpreting it directly as a ``likelihood,'' or by using it as a measurement of typicality.  We found in our experiments that tests solely utilizing the likelihood of a deep generative model were often vulnerable to OOD data.  \cite{nalisnick2019detecting} attributed this to a defect in deep generative models themselves.  In this section, we aim to show that this is at least partially due to the methodologies for OOD detection rather than pathologies of generative models themselves.

Let us consider a flow-based model, such as Glow \citep{kingma2018glow}.  In flow-based models, the log-likelihood is computed as $\log{q(X|\theta_n)}= T^{(\text{latent})}_{n}(X) + T^{(\text{jac})}_{n}(X)$, where $T^{(\text{jac})}_{n}(X)\defeq\log |\J(X)|$ is the Jacobian of the transformation from $X$ to $Z$, and $T^{(\text{latent})}_{n}(X)=q(Z|X,\theta_n)$ is the log-probability of the latent variable $Z$.  Consider the example shown in Figure~\ref{fig:metric_space}.  In this example, we show the two-dimensional distribution of metrics for an in-distribution dataset (blue) and an OOD dataset (red).  The two dimensions in this case are $q(Z)$ and $\log \left|\J\right|$, which are added together to compute the log-likelihood.  From this, it is straightforward that curves of constant likelihood have a slope of -1 in this space.  

Consider how different decision rules reject data in this space.  If we assume that data with low likelihood were OOD, then our decision rule would be approximately equivalent to that shown in the left panel of Figure~\ref{fig:metric_space}.  If we instead use the typicality test (TT) from \cite{nalisnick2019detecting}, we observe the result shown in the center panel.  Effectively, excluding examples with low log-likelihood determines a half-space for which the data is assumed to be in-distribution.  Similarly, TT identifies in-distribution data as the intersection of two half-spaces.  However, in both cases, OOD data falls within the region classified as in-distribution.  As a result, both metrics do extremely poorly on OOD detection.  In contrast, DoSE operates over each dimension of the space separately (or all jointly), and is able to find a more optimal decision boundary.

This behavior is not restricted to flow-based models.  In VAEs, the log-evidence is approximated as $q(X)=\E_{Z\sim q(Z|X)}\left[q(X|Z)r(Z)/q(Z|X)\right]$, a nonlinear function of a sum of the cross-entropy between the posterior and the prior, the log-likelihood from the decoder, and the entropy of the encoder.  Therefore, models that use only the log-evidence of a VAE as a decision rule can exhibit a similar vulnerability to a flow-based model doing the same.

Furthermore, we observe this phenomenon experimentally.  \autoref{fig:statistic_space} shows a decomposition similar to \autoref{fig:metric_space} for a model trained on CelebA, using CIFAR10 as OOD data.  We observe that, in this space, the OOD data is projected such that it is nearly perfectly confounded for both $q(X|\theta_n)$ as well as TT.  DoSE operates on the granularity of the statistics themselves, and therefore achieves a much better AUROC because it partitions the space using all of the constituent statistics, from which the OOD data is noticeably shifted from the in-distribution data.

\section{Additional details of experiments}
\label{appendix:experiments}

To evaluate the performance of DoSE, we first train a generative model on an in-distribution dataset, and fit density of states estimators to statistics from the generative model on the training set.  Before performing inference, we evaluate the memorization of the model using a random heldout set of 10\% of the training examples.  When performing inference, we compute the same set of statistics from the generative model on new input data, and calculate the DoSE scores for each example.  We measure performance use the AUROC measured using the DoSE scores found from an evaluation set and a specific OOD set.  For each trained model, we evaluate the performance against multiple OOD datasets. 

\textbf{Datasets.} We use common dataset pairings for the OOD detection task. For our in-distribution datasets, we use MNIST and Fashion MNIST, along with CIFAR10, Street View Housing Numbers (SVHN), and CelebA.  These datasets are then paired with the other datasets having the same dimensions, which are taken to be OOD data.  Similar to \citep{choi2018waic, devries2018learning}, we also use uniform and Gaussian noise, and horizontally- and vertically-flipped versions of the in-distribution test set as additional OOD datasets. We also use Omniglot for MNIST and Fashion MNIST, and ImageNet-32 and CIFAR100 for CIFAR10, SVHN, and CelebA.  

Many of these dataset pairings are ``simple,'' in that likelihood alone would be a reasonable rule to detect OOD data.  However, there are several ``hard'' OOD dataset pairings identified by previous work.  FashionMNIST$\rightarrow$MNIST and CIFAR10$\rightarrow$SVHN were both identified as difficult dataset pairings by \cite{nalisnick2018do}.  Additionally, \cite{nalisnick2019detecting} identified CelebA$\rightarrow$ CIFAR10/100 and CIFAR10$\rightarrow$CIFAR100 to be particularly difficult pairings.  The latter is particularly difficult, since both are subsets of the 80 million tiny images dataset \citep{torralba200880}, but have non-overlapping class labels.

\textbf{Architectures.} Similar to \citep{choi2018waic, nalisnick2019detecting}, we train $\beta$-VAEs \citep{higgins2017beta} for MNIST and Fashion MNIST, and Glow \citep{kingma2018glow} for CIFAR10 and SVHN. For the $\beta$-VAE models, our encoder and decoder followed the architecture from \cite{choi2018waic}.  We use a 2-dimensional latent space, and a trainable mixture of 200 Gaussians for the marginal distribution $r(Z)$.  We also considered higher dimensional latent spaces where the model would measure higher log-likelihoods, and found that the DoSE results were similar, but the results from competing techniques worsened substantially. We fix the mean and logit of the first component of $r(Z)$ to improve training stability.  For MNIST, we use a Bernoulli distribution for the decoder log-likelihood.  For FashionMNIST, we instead used a Logit-Normal distribution, a bijective transformation of the normal distribution to the interval $(0, 1)$ using a sigmoid bijector, since the majority of the spatial variation between pixels in FashionMNIST occurs at values near 0.5, where the Bernoulli distribution struggles to capture variation.

For the Glow models, we replicated the architecture from \citep{nalisnick2019detecting}, using 3 spatial hierarchies of 8 steps of the flow.  Each step of the flow consists of \textit{actnorm}, an invertible $1 \times 1$ convolution, and an affine coupling layer. We use a RealNVP bijector \citep{dinh2016density} for the coupling layer, which uses a 3-layer convolutional stack with ReLU activations and 400 filters.  For stability in training, the last convolutional layer is set to 0 at initialization for each stack, which corresponds to the full Glow network simply producing an identity transformation (with some rearranging of the pixels) at initialization.  Between each spatial hierarchy, we remove half of the data to create multiple different levels of spatial variation.  Altogether the Glow network constructs a bijective transformation which projects the data $X$ into a latent space $Z$ with the same dimensionality (3072, in these experiments).  The full Glow model is then created as a transformed distribution using $\mathcal{N}(0,1)$ as the base distribution, and the Glow network as the bijector.   All experiments were performed using TensorFlow and TensorFlow Probability \citep{tensorflow2015-whitepaper}. 

\textbf{Training details} 
Following \citet{kingma2018glow}, we train Glow models using the Adamax optimizer \citep{kingma2014adam} with a learning rate initialized to 0 and gradually increased to 0.001 over 10 epochs, after which point it is held constant. We trained the models for 250 epochs in total.  We optimize the negative log-likelihood $q(X|\theta_n)$ with added $L_2$-regularization of the weights to reduce memorization in the model.  We explored regularization constants of $\lambda=[0., 0.01, 0.05, 0.1, 0.5]$, and determined that $\lambda=[0.05, 0.1]$ limited memorization without also limiting generative model performance.

For VAE models, convergence was much faster, so we train for 50 epochs using a learning rate initialized at 0.0001, and decayed exponentially by half every 10000 training steps.  We follow \cite{choi2018waic} and use the Adam optimizer to optimize the traditional Evidence Lower Bound (ELBO).  We evaluate the ELBO using 16 samples from the posterior distribution.  To prevent memorization, we employ two additional procedures:  First, we ``burn-in'' the decoder for one epoch by drawing samples from the prior, and use the decoder to estimate the log-likelihood for each input given the samples.  This has the effect of initializing our likelihood to be properly conditioned on the prior, keeping small initial gradients for the encoder early on in training.  Second, we employ ``reverse beta-annealing'' during training.  We start with a large value of $\beta=100$, and we decay its value by a factor of 2.0 every 3 epochs.  We found that this causes the posterior to be more effectively anchored to the prior during training, which ultimately results in more informative latent spaces and a more useful sampling distribution (and therefore more reliable outlier detection).

For each dataset, we trained 5 separate models following \citep{lakshminarayanan2017simple, choi2018waic, nalisnick2019detecting}. This allows us to both quantify the variability in performance over separate training runs, as well as to utilize an ensemble of all 5 models in order to produce a stronger and more robust estimator. 

\textbf{Evaluation of performance.} Once a model is trained, we construct our DoSE by measuring the value of summary statistics of the model, computed on the elements of the training set.  For VAEs, we have an abundance of possibilities: 
\begin{itemize}
\item KL divergence between the posterior and marginal $T^{(\text{rate})}_{n}(X)=\KL[q(Z|X,\theta_n), q(Z)]$ (rate)
\item Cross-entropy between the posterior and marginal $T^{(\text{xent})}_{n}(X)=\H[q(Z|X,\theta_n), q(Z)]$
\item Entropy of the posterior $T^{(\text{ent})}_{n}(X)=\H[q(Z|X,\theta_n)]$
\item Expected log-likelihood computed over the posterior $T^{(\text{dist})}_{n}(X)=\E_{q(Z|X,\theta_n)}[q(X|Z,\theta_n)]$ (distortion)
\item Estimate of the evidence computed using a 16-sample importance weighted autoencoder (IWAE) given by $T^{(\text{iwae})}_{n}(X)=q(X|\theta_n) = \E_{q(Z|X)}[q(X|Z,\theta_n)q(Z)/q(Z|X,\theta_n)]$ (log-likelihood, following the terminology of \cite{nalisnick2018do,nalisnick2019detecting,choi2018waic,ren2019likelihood})

\end{itemize}

For Glow models, the number of statistics is more constrained because Glow does not have as many ways to evaluate summaries on the generative model.  In this work, we use:
  \begin{itemize}
      \item ``Log-likelihood'' $T^{(\text{like})}_{n}(X)=q(X|\theta_n)$, and its two constituents
      \item Log-probability of the latent variable $T^{(\text{latent})}_{n}(X)=q(Z(X)|\theta_n)$
      \item Log of the determinant of the Jacobian between $X$ and $Z$ (i.e., $T^{(\text{jacobian})}_{n}(X)=\log |\J(X)|$)
  \end{itemize}


For each statistic that we measure in the training set, we compute a Kernel Density Estimate (KDE), using the default implementation in SciPy \citep{2020SciPy-NMeth} to build an individual DoSE.  $DoSE_{KDE}$ is then simply the sum over all the DoSE scores for an individual statistic:
\begin{equation}
    DoSE_{KDE} = \sum_{j}^{m}KDE_{j}(x) = \sum{j}^m \frac{1}{nh} \sum_i^n \phi\left(\frac{T_j(x) - T_j(x_i)}{h}\right)
\end{equation}

We build $DoSE_{SVM}$ by creating an $n$-dimensional feature vector of the $n$ metrics for each observation.  We first use Principal Components Analysis (PCA) to learn a whitening transformation from the training set to help correct against the wildly different variance observed in different statistics.  We then use the transformed space to learn a one-class SVM.  Both PCA and the SVM use the default implementations in scikit-learn \citep{scikit-learn}.  

Before we evaluate the DoSE performance on OOD data we check its memorization.  To do this we measure the expected calibration error (ECE) \citep{guo2017calibration} of $DoSE_{KDE}$ using a small heldout subset of 10\% of the examples from the training set.  These examples are in-distribution but never seen during training, and therefore the ECE measures the degree to which the DoSE scores given to new in-distribution data are consistent with the scores given to data seen during training.  In our experiments, we found that without some form of intervention, both VAE and Glow models exhibited extreme capacity for memorization, and therefore had high ECE.  This inspired our earlier described preventative measures, such as reverse beta-annealing for VAEs, and $L_2$-regularization for Glow.  Using these additional procedures, we found that our memorization scores were typically around 1\% for most models.

We evaluate the performance of $DoSE_{KDE}$ and $DoSE_{SVM}$ by computing the scores on the specified OOD datasets, and use these scores to measure the AUROC for OOD detection.  We compare our method against four unsupervised baselines: the vanilla likelihood $q(X|\theta_n)$, Watanabe-Akaike information criterion (WAIC) \citep{choi2018waic}, the typicality test (TT) using a batch size of 1 (which represents a more realistic application than a larger batch size), \citep{nalisnick2019detecting}, and likelihood ratios (LLR) \citep{ren2019likelihood}. For WAIC, we use Eq. 1 from their paper to compute the scores. For LLR, we train a background model using their method of mutations to perturb the input data. We use a mutation rate of 0.15, which is in the middle of the range of values they found to produce acceptable results.  The LLR score is then the difference between the scores from models trained without and with mutations. With the exception of the background models used for LLR, all methods are evaluated on the same models.  This provides an apples-to-apples comparison between methods, and highlights the differences between them as a function of the method itself, rather than the underlying model.

\section{Degradation of Signal Due to Uninformative Statistics}\label{appendix:degradation}

\begin{figure}
    \centering
    \includegraphics[width=0.5\hsize]{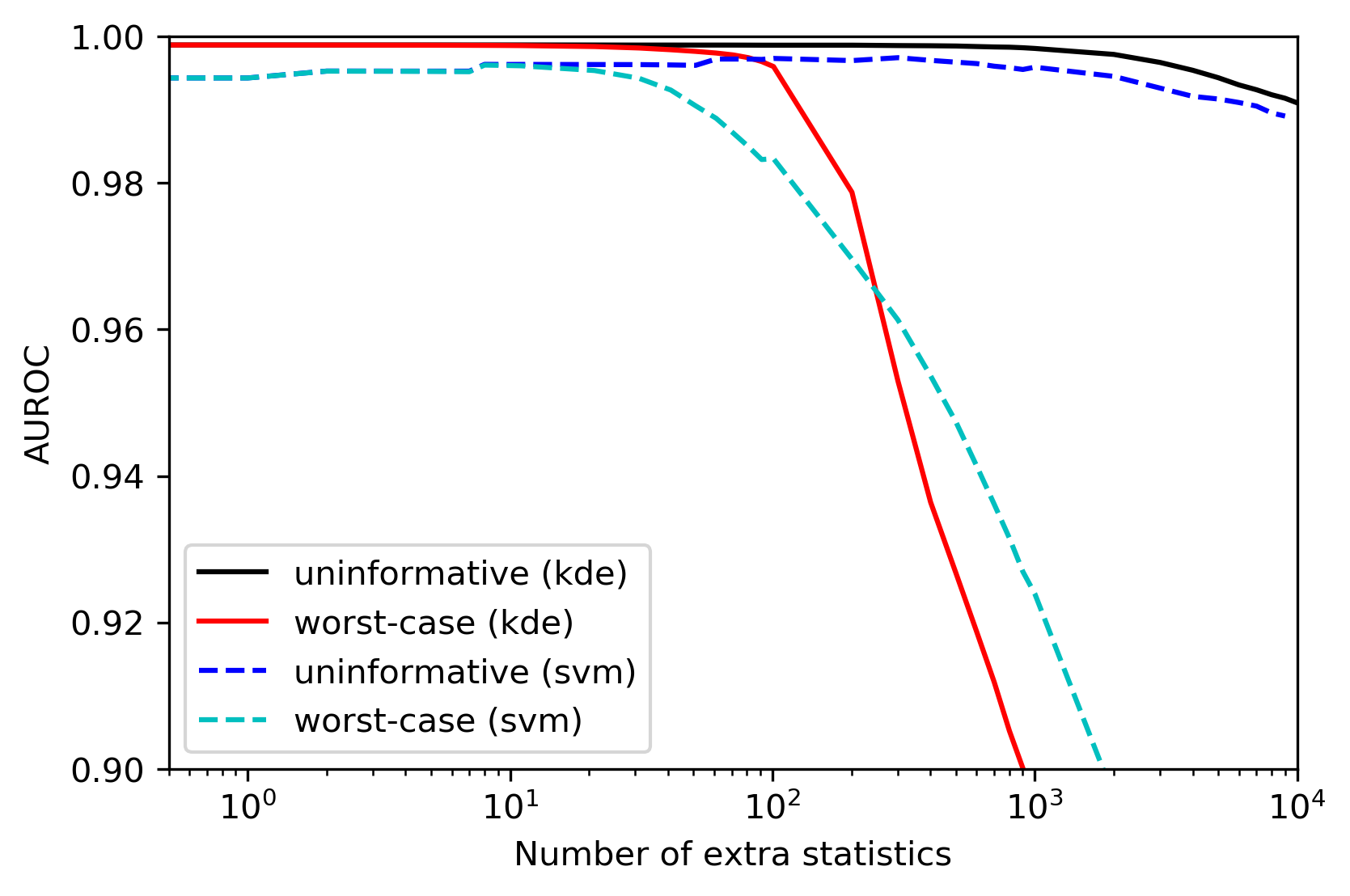}
\caption{The AUROC observed when using DoSE on FashionMNIST$ \rightarrow $MNIST with the addition of extra statistics, which are either uninformative, or purposely obfuscatory.  We find that uninformative statistics have little effect on performance, with only a $1\%$ drop in the AUROC even after the addition of $10^4$ uninformative statistics.  Performance degrades by the same amount when roughly 80 obfuscatory statistics are used.}
    \label{fig:superfluous_statistic}
\end{figure}

In our experiments, we used statistics which were useful diagnostics of the model performance, and which we therefore expected to contain some degree of meaningful signal about whether an input was in-distribution or OOD.  When deploying DoSE on different types of models, one may not always have access to the same statistics or be tempted to choose as many statistics as are available.  As we showed in \Cref{sec:intro}, certain statistics are not be able to identify certain OOD datasets as atypical.  A question we wanted to probe experimentally was then; ``How hazardous are uninformative statistics for the OOD signal?''  Since we do not have access to OOD data during training, answering this question will allow us to be slightly more liberal with choosing statistics.

For this experiment, we chose to use the FashionMNIST$\rightarrow$MNIST pairing.  We took the  $DoSE_{KDE}$ scores evaluated on the FashionMNIST and MNIST test sets.  We then added ``superfluous'' DoSE scores.  To do this, we assumed that a new statistic was evaluated, given by $T^{(\text{useless})}\sim \mathcal{N}(0,1)$, which was distributed identically for both the in-distribution data and the OOD data.  The DoSE score for this superfluous statistic is then simply the log-likelihood: $\log q(T^{(\text{useless})}) = -0.5 T^{\text{useless}~ 2} -\log{\sqrt{2\pi}}$.  We repeatedly drew more of these useless statistics, and added them to the test and OOD DoSE scores.  We also further consider a worst-case statistic, for which OOD data is given maximally typical scores ($-\log \sqrt{2\pi}$ for the unit-normal distribution).

We show the AUROC as a function of the number of superfluous dimensions in \autoref{fig:superfluous_statistic}.  For this dataset pairing, we find that even after an extremely large number of superfluous statistics (at least $3\times10^{5}$), the AUROC has only decayed by 0.04, and would therefore still have higher AUROC on this dataset pairing than any competing technique.  This phenomenon is observed using both a KDE and a SVM to evaluate the DoSE scores.  In the worst case scenario, as expected we find that the number of statistics needed to degrade the OOD signal is much smaller, requiring only 100 statistics to produce noticeable degradation for the KDE, and only roughly 20 for the SVM.  Even here, we find that roughly 300 statistics are necessary to drop the DoSE performance below alternative methods.   Empirically this suggests that there may not be a strong need to carefully choose statistics.

\clearpage
\newpage
\section{Histograms of Statistics}
\label{appendix:histograms}

\begin{figure}[!htb]
    \centering
    \includegraphics[width=\hsize]{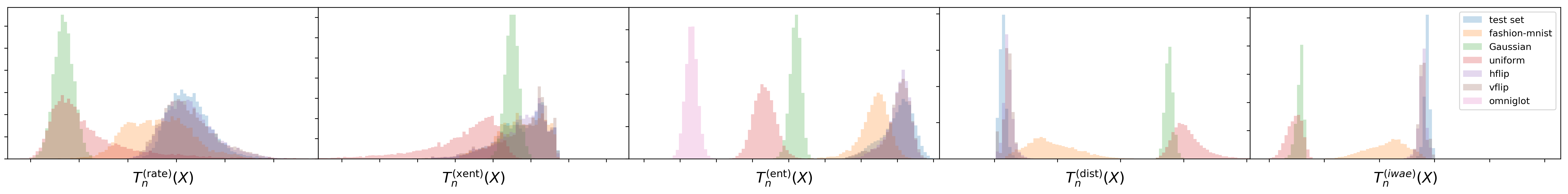}
    \caption{Histograms of 5 different statistics evaluated on a VAE trained on the MNIST dataset.  The leftmost column shows the KL divergence between the posterior and the prior.  The second column shows the cross-entropy between the posterior and the prior.  The third column shows the entropy of the encoder.  The fourth shows the distortion (the expected log-likelihood from the decoder).  The last column shows the log-evidence, computed using a 16-sample IWAE estimate.  For each metric, we show the distribution of that metric observed in the test set, along with multiple different OOD datasets.}
    \label{fig:MNIST_metrics}
\end{figure}

\begin{figure}[!htb]
    \centering
    \includegraphics[width=\hsize]{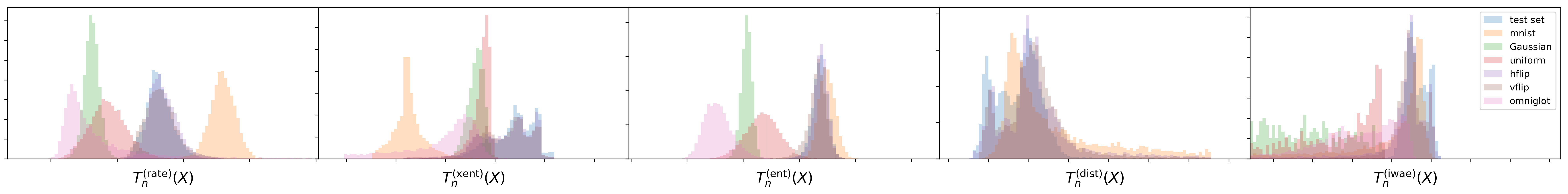}
    \caption{Same as Figure~\ref{fig:MNIST_metrics}, but for a VAE trained on FashionMNIST.  Note that while the log-likelihood is a successful OOD detection metric when trained on MNIST, it does not perform similarly when trained on FashionMNIST, often overlapping strongly with various OOD datasets.  Other statistics, such as the KL divergence between the posterior and the prior appear to be much more informative in this case.}
    \label{fig:Fashion_metrics}
\end{figure}

\begin{figure}[!htb]
    \centering
    \includegraphics[width=\hsize]{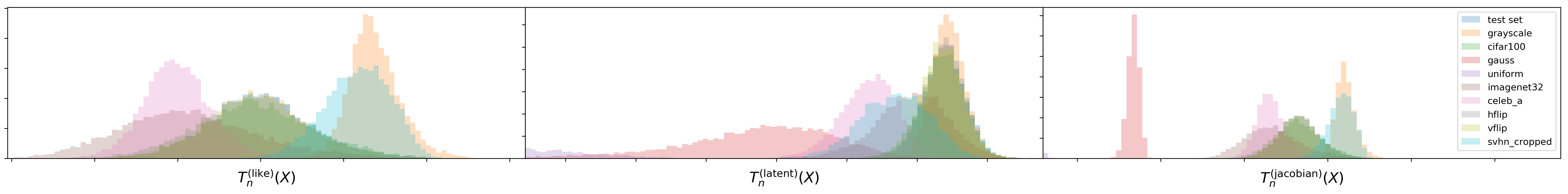}
    \caption{Same as Figure~\ref{fig:MNIST_metrics}, but for a Glow model trained on CIFAR10.}
    \label{fig:Cifar_metrics}
\end{figure}

\begin{figure}[!htb]
    \centering
    \includegraphics[width=\hsize]{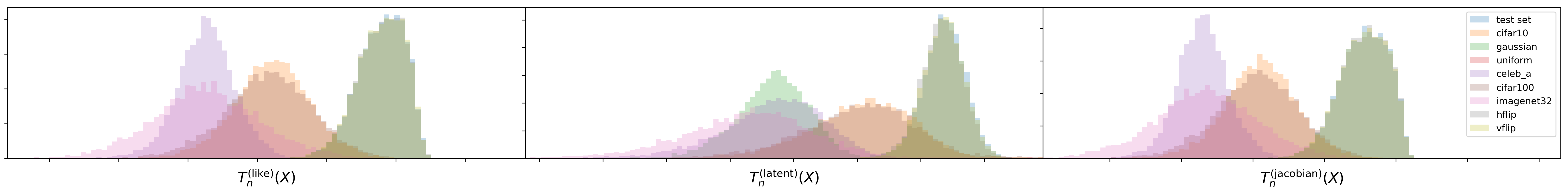}
    \caption{Same as Figure~\ref{fig:MNIST_metrics}, but for a Glow model trained on SVHN.}
    \label{fig:SVHN_metrics}
\end{figure}

\begin{figure}[!htb]
    \centering
    \includegraphics[width=\hsize]{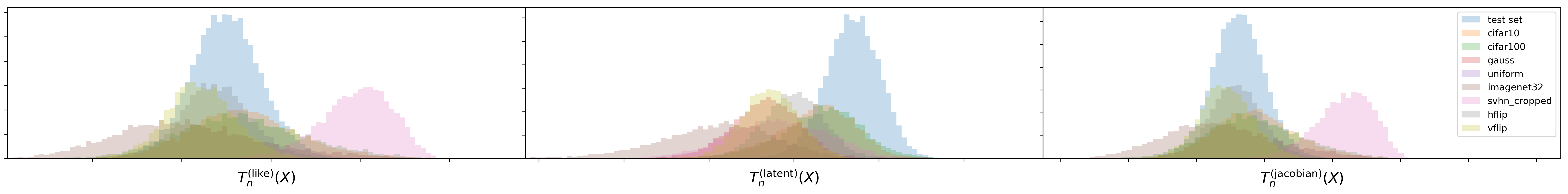}
    \caption{Same as Figure~\ref{fig:MNIST_metrics}, but for a Glow model trained on CelebA.}
    \label{fig:CelebA_metrics}
\end{figure}

\end{document}